\documentclass[10pt,twocolumn,letterpaper]{article}
\usepackage[pagenumbers]{cvpr}

\usepackage{amsmath,amsfonts,bm}

\def\eqref#1{equation~\ref{#1}}

\def\1{\bm{1}}

\DeclareMathAlphabet{\mathsfit}{\encodingdefault}{\sfdefault}{m}{sl}
\SetMathAlphabet{\mathsfit}{bold}{\encodingdefault}{\sfdefault}{bx}{n}

\usepackage{graphicx}
\graphicspath{{figs/}}
\usepackage{multirow}
\usepackage{xcolor}
\usepackage{booktabs}
\definecolor{cvprblue}{rgb}{0.21,0.49,0.74}
\usepackage[pagebackref,breaklinks,colorlinks,allcolors=cvprblue]{hyperref}
\usepackage{url}
\usepackage{amsthm}
\usepackage[inline]{enumitem}
\usepackage[table]{xcolor}  
\definecolor{bestgray}{gray}{0.88}  
\newcommand{\best}[1]{\cellcolor{bestgray}#1}
\definecolor{deltagreen}{RGB}{46,125,50}
\definecolor{deltared}{RGB}{183,28,28} 
\newcommand{\up}[1]{\,{\scriptsize\textcolor{deltagreen}{($\uparrow$#1)}}}
\newcommand{\down}[1]{\,{\scriptsize\textcolor{deltared}{($\downarrow$#1)}}}

\newtheorem{theorem}{Theorem}[section]
\newtheorem{lemma}[theorem]{Lemma}
\newtheorem{corollary}[theorem]{Corollary}
\theoremstyle{definition}
\newtheorem{definition}{Definition}[section]

\newcommand{\method}{MotionCore}

\title{MotionCore: A Latent Neural Simulator for Video Generation with Rigid-Body Dynamics}

\author{Zheng Wang\textsuperscript{3,1}\thanks{Work done during an internship at TeleAI}  , Zhengwei Gong\textsuperscript{3,2}\footnotemark[1] , Yuanzhi Liang\textsuperscript{3}, Haibin Huang\textsuperscript{3}, Chi Zhang\textsuperscript{3} \thanks{Corresponding to \href{mailto:zhangc120@chinatelecom.cn}{zhangc120@chinatelecom.cn}} \\
\textsuperscript{1}  The University of Tokyo \\
\textsuperscript{2}  Shanghai Jiao Tong University  \\
\textsuperscript{3}  Institute of Artificial Intelligence, China Telecom (TeleAI) \\ }

\def\paperID{*****}
\def\confName{CVPR}
\def\confYear{2026}

\begin{document}

\maketitle

\begin{abstract}

    Video generation models can produce highly realistic videos while generating physically inconsistent object motion. This limitation suggests that visual modeling alone may not provide a transferable representation of physical dynamics. We introduce \method{}, a framework for learning physical dynamics via a learnable latent neural operator which focuses on rigid-body video generation. We introduce a latent neural simulator learning to predict the evolution of object-level physical states from observed motion. Its predictions are then converted into spatially aligned latent controls for a controllable video diffusion model. The generator serves as a visual interface: it provides appearance and scene detail, while the learned dynamics model specifies the physical trajectory. We construct our training dataset comprising semi-realistic synthetic multi-object videos with rich annotations. We evaluate \method{} on public physical video generation benchmarks and conduct controlled ablations to assess the contribution of learned dynamics relative to other priors. Results indicate that \method{} improves physical consistency and trajectory obedience in the evaluated rigid-body settings, while maintaining comparable visual quality. \method{} outperforms larger general and physics-aware video generators while using substantially fewer parameters.

\end{abstract}

\section{Introduction}

\begin{figure}[t]
    \centering
    \includegraphics[width=\linewidth]{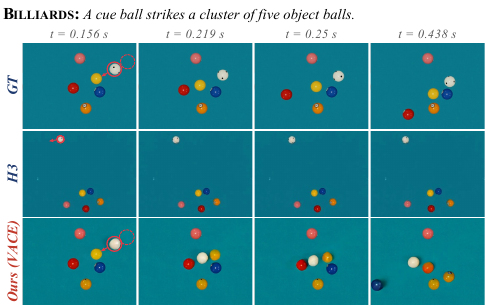}
    \caption{\textbf{\method{} on billiards.} Rows show the ground truth, the depth condition rendered from the LNO-predicted object states, and the video generated by the frozen VACE-1.3B from that condition. Circles and arrows mark the balls involved in the collisions.}
    \label{fig:teaser}
\end{figure}

Video generation models have made rapid progress in visual quality. They can synthesize detailed scenes, realistic appearances, and complex motions. Yet visual realism does not imply physical correctness. Generated objects may accelerate without a clear cause, pass through one another, or follow trajectories that conflict with collisions and gravity. This limitation not only occurs among older or small-scale models, but also among video foundation models such as Seedance 2.0~\citep{seedance2026seedance2} and MiniMax H3~\citep{minimax2026h3}. Recent benchmarks, including VideoPhy, PhyGenBench, Physics-IQ, and VideoPhy-2, report a persistent gap between visual quality and physical consistency \citep{bansal2025videophy,meng2024phygenbench,motamed2026physics,bansal2026videophy2}. These observations suggest that a strong visual prior alone is insufficient for reliable physical prediction. More broadly, improving visual modeling capacity does not automatically provide a transferable representation of object dynamics.

Although generating a physically plausible video is a challenging task, videos themselves are a natural source of physical observations. They contain object trajectories, interactions, and changes in motion under diverse conditions. Object-level representations can be inferred from video and used for learned dynamics prediction~\citep{watters2017visual}. This suggests a modular view of video generation (Fig.~\ref{fig:teaser}). 
A physical dynamics module learns how the observed system evolves, and a pretrained video diffusion model then acts as a visual tool that renders the predicted state with realistic appearance. In this setting, video generation works like a piloted mech: the dynamics-learning part is the pilot that decides how the scene moves next, and the diffusion model is the body that carries out that decision by rendering it into frames.

However, how to learn physical dynamics beyond visual cues becomes a major obstacle. Physical dynamics can typically be described by differential equations with unknown parameters: rigid-body motion is governed by the Newton–Euler equations, and the propagation of light by Maxwell's equations. This suggests that a solver tailored to a specific scene can serve as an abstract description of that real-world scene. We learn such a solver in latent space with a neural operator, and use it to build a path from physical space to visual space. 

In this paper, we introduce \method{}, a framework for video-grounded rigid-body dynamics learning. \method{} uses a latent neural operator~\citep{wang2024latent} to model the evolution of object-level physical states. The neural operator predicts future object states from observed motion and produces intermediate latent features that describe the evolving physical system. And we use a controllable video generation model VACE~\citep{jiang2025vace} built on Wan2.1-1.3B~\citep{wan2025} as the video generator. The generator remains frozen and provides appearance, texture, and visual detail, while the dynamics model provides the physical trajectory. Our main contributions are as follows:
\begin{itemize}
    \item We model the rigid-body dynamics prediction as solution to partial differential equations by treating rigid bodies with high stiffness and therefore leveraging the generalization capabilities of neural operators to deal with complex environment settings.
    \item We give the formal proof of theoretical approximation error boundary for latent neural operators, which motivates us to control the error via choosing geometry-based latent as intermediate representation.
    \item \method{} outperforms larger video diffusion models and various physics-aware video generators with substantially fewer trainable parameters, demonstrating the effectiveness of our approach.
\end{itemize}

\section{Related Work}

\paragraph{Neural operators in latent space.}
Neural operators learn maps between function spaces and can therefore transfer across spatial discretizations. DeepONet represents nonlinear operators with branch and trunk networks~\cite{lu2021learning}, whereas the Fourier Neural Operator performs global kernel integration in the spectral domain~\cite{li2021fourier}. Latent formulations reduce the cost of evolving high-dimensional physical fields by learning a compact solution manifold or an encoder--operator--decoder factorization. NOMAD uses a nonlinear decoder for finite-dimensional solution manifolds~\cite{seidman2022nomad}. Latent Spectral Models and the Latent Neural Operator learn the principal transformation in a compact space~\cite{wu2023latent,wang2024latent}, while latent DeepONet compresses both input and output functions before operator learning~\cite{kontolati2024learning}. Geometry-aware variants map irregular meshes or particles to latent grids or tokens~\cite{li2023geometry,alkin2024universal}, and AROMA retains spatially organized latent sequences and learns their temporal evolution with a conditional Transformer and a diffusion objective~\cite{serrano2024aroma}. More recent models use state-space dynamics for long-range propagation~\cite{tiwari2025latent} or flow-matching autoregression for generative PDE evolution~\cite{koupai2025enma}. AROMA and ENMA are especially relevant because they combine compressed physical states with generative dynamics, but their outputs remain numerical PDE fields. \method{} instead places a PDE-informed solution operator inside video latent denoising, using the operator state to guide the generation of detailed visual frames.

\paragraph{Simulator-based physical video generation.}
One line of work obtains motion from an explicit simulator and transfers it to a generative visual model. MotionCraft warps diffusion noise latents with optical flow computed from a physical simulation~\cite{montanaro2024motioncraft}. PhysGen infers scene properties, simulates rigid-body motion, and renders the result with video diffusion~\cite{liu2024physgen}, while PSIVG reconstructs a dynamic scene and places a simulator in the generation loop~\cite{foo2026psivg}. These pipelines provide interpretable control, but their behavior is coupled to reconstructed object states, simulator assumptions, or test-time optimization. A complementary line learns physical control or supervision from data. PISA combines simulated-video fine-tuning with reward-based post-training for free fall~\cite{li2025pisa}. PointVid regularizes generation with 3D point trajectories~\cite{chen2025pointvid}, Force Prompting conditions videos on local or global force fields~\cite{gillman2025force}, and VLIPP converts vision-language reasoning into coarse motion plans~\cite{yang2025vlipp}. VideoREPA instead distills pairwise spatiotemporal relations from a video representation model~\cite{zhang2025videorepa}. These methods inject useful motion variables, plans, or preferences, but do not learn the governing field evolution as a reusable solution operator.


\section{Preliminaries}
\label{sec:preliminaries}

The trajectory of a rigid body is the solution of its governing equations of motion. In Sec.~\ref{sec:prelim-representation}, we formulate the rigid-body dynamics as a set of partial differential equations instead of Newton-Euler equations, which covers free fall, sliding, collisions, swinging and so on. Then in Sec.~\ref{sec:prelim-neural-operators}, we briefly give the definition and properties of neural operators.

\subsection{Representation of Rigid-body Dynamics}
\label{sec:prelim-representation}
A physical trajectory records one realization of a dynamical system, whereas its governing equation describes an entire family of realizations. The governing equations of rigid-body dynamics are usually in the forms of ordinary differential equations (ODEs)~\citep{marsden1999rigiddynamics}. For example, in the case of motion without collisions and external constraints, the Newton-Euler equations are given by Eq.~\eqref{eq:newton-euler}.
\begin{equation}
    \label{eq:newton-euler}
    \dot{\mathbf{p}} = \mathbf{v}, m\dot{\mathbf{v}} = \mathbf{F},
    \dot{\mathbf{R}} = \mathbf{R}\widehat{\boldsymbol{\omega}},
    \mathbf{I}\dot{\boldsymbol{\omega}} + \boldsymbol{\omega}\times
    \left(\mathbf{I}\boldsymbol{\omega}\right) = \boldsymbol{\tau}
\end{equation}
where $\mathbf{p}, \mathbf{v}, \mathbf{F}, \mathbf{I}, \boldsymbol{\tau}, \mathbf{R}, \boldsymbol{\omega}$ specify the position , velocity, net external force, inertia tensor. net external torque, rotation matrix and angular velocity of the rigid body's mass center. If $\mathbf{F}$ and $\boldsymbol{\tau}$ are uniquely determined by current states and time, the govern equations could be reduced to one ODE which takes the form
\begin{equation}
    \dot{\mathbf{x}} = f(t, \mathbf{x}; \boldsymbol{\phi})
    \label{eq:diff-equation}
\end{equation}
where $\boldsymbol{\phi}$ includes settings mentioned in Eq.~\eqref{eq:newton-euler}. Eq.~\eqref{eq:diff-equation} is a finite-dimensional ODE evolving on a physical-state manifold. However, not all rigid-body motion can be described by this smooth ODE such as impulsive dynamics in the scene of ideal rigid-body collision and externally constrained rigid-body system. 

To address this limitation, we approximate rigid bodies as deformable elastic body with large but finite stiffness, and collisions through elastic-body's contact laws~\cite{paoliI, paoliII}. The observed instances are modeled as elastic continua, which means we could describe general rigid-body dynamics via partial differential equations (PDEs), unifying smooth and impulsive response. Our aim is to find an intermediate physical-state representation which can guide the video diffusion models generate plausible outputs. 

\subsection{Neural Operator}
\label{sec:prelim-neural-operators}


Neural Operators have following properties that enable it to effectively learn mappings between functions~\citep{lu2021learning,li2021fourier}. 
\begin{enumerate*}[label=(\roman*)]
    \item Initial conditions, coefficients, and boundary data enter as inputs rather than as fixed settings, so one set of parameters represents a whole family of solutions. 
    \item The same parameters apply across compatible discretizations: inputs can be sampled on different grids, outputs can be queried at any resolution, and predictions converge as the discretization is refined. This transfer concerns different samplings of the same instance, and extrapolation to unseen physical regimes remains a separate question. 
    \item A trained NO evaluates a new instance in one forward pass, amortizing the cost of a numerical solve.
\end{enumerate*}
Latent neural operators~\citep{wang2024latent} reduce cost of full-resolution evolution by factorizing the operator into latent space. Section~\ref{sec:physical-operator} adopts this factorization for structured physical states and analyzes what the latent representation must preserve for the approximation to hold.
\providecommand{\brain}{\mathcal{L}}
\providecommand{\state}{\mathbf{s}}
\providecommand{\pstate}{\hat{\mathbf{s}}}
\providecommand{\proj}{\pi}

\section{Theoretical Bound}
\label{sec:physical-operator}

\subsection{Problem Setting}
\label{sec:problem-setting}

We study rigid-body video generation with dynamics prediction separated from visual rendering. A semantic condition, typically a text prompt $\mathbf{T}$ with the first frame $\mathbf{I}_{\mathrm{ref}}$, specifies visual appearance and initial conditions. 

The envisioned model should receive the geometry and object states of the initial frame, and predict future trajectories at targeted video timestamps $1,\dots,K$, then a diffusion model renders trajectories into realistic videos. However, there are some particular obstacles:
\begin{itemize}
    \item \textbf{Limited Access to 3D Assets.} Current 3D reconstruction methods mainly depends on multi-view observations~\citep{2016mvf,mildenhall2020nerf,kerbl20233dgs} which are usually valuable.
    \item \textbf{Hybrid and Complicated System.} In real dynamic scenes, there exists non-smooth dynamics like collision. And different structural constraints could lead to distinct motion patterns. 
\end{itemize}

Following Section~\ref{sec:preliminaries}, a scene contains $K$ bodies, and body $k$ has state $\mathbf{x}_k=(\mathbf{p}_k,\mathbf{R}_k,\mathbf{v}_k,\boldsymbol{\omega}_k)$. Between contacts, $\mathbf{x}_k$ follows Eq.~\eqref{eq:newton-euler}. At contact this description breaks down, so we adopt the compliant model of Section~\ref{sec:preliminaries}: each body is treated as an elastic body with large but finite stiffness, and contact follows elastic contact laws~\cite{paoliI,paoliII}. The coefficients $\boldsymbol{\phi}\in\Phi$ collect mass, inertia, stiffness, and contact parameters. Under this model the trajectory remains continuous in time through collisions, and the impulsive rigid-body motion is recovered as the stiffness tends to infinity~\cite{paoliI,paoliII}. With geometry and $\boldsymbol{\phi}$ fixed over the horizon $[0,T_h]$, the dynamics define the solution operator
\begin{equation}
    \mathcal{S}^{\dagger}:
    \mathcal{X}^{K}\times\Phi \longrightarrow
    C\big([0,T];\mathcal{X}^{(K)}\big),
    (\mathbf{x}(0),\boldsymbol{\phi})\longmapsto\mathbf{x}(\cdot)
    \label{eq:rigid-body-solution-operator}
\end{equation}
where $\mathcal{X}$ is the state space of single rigid body. This is the coefficient-to-solution setting of operator learning in Section~\ref{sec:prelim-neural-operators}. Just as a permeability field determines the pressure of a Darcy flow~\citep{li2021fourier}, the coefficient field and the initial state determine a rigid-body trajectory, and one operator covers the entire family of scenes. Learning $\mathcal{S}^{\dagger}$ therefore means learning the shared law of rigid-body motion rather than any single trajectory.

We learn this operator on an encoded state. The observed history provides the initial state. An encoder $\mathcal{E}$ maps the history to latent codes $\mathcal{Z}$, a neural operator $\mathcal{T}_N$ evolves the code, and a decoder $\mathcal{D}$ returns the predicted trajectory~\citep{wu2023latent,wang2024latent}. Because the token has a fixed size independent of the number and shape of objects, one operator can serve environments that an explicit simulator would have to set up one by one. Theorem~\ref{thm:lno_bound} makes this precise: the error is controlled by how well the token preserves the evolution and by how simple the induced latent operator is.

\begin{figure*}[tb]
  \centering
  \includegraphics[width=\linewidth]{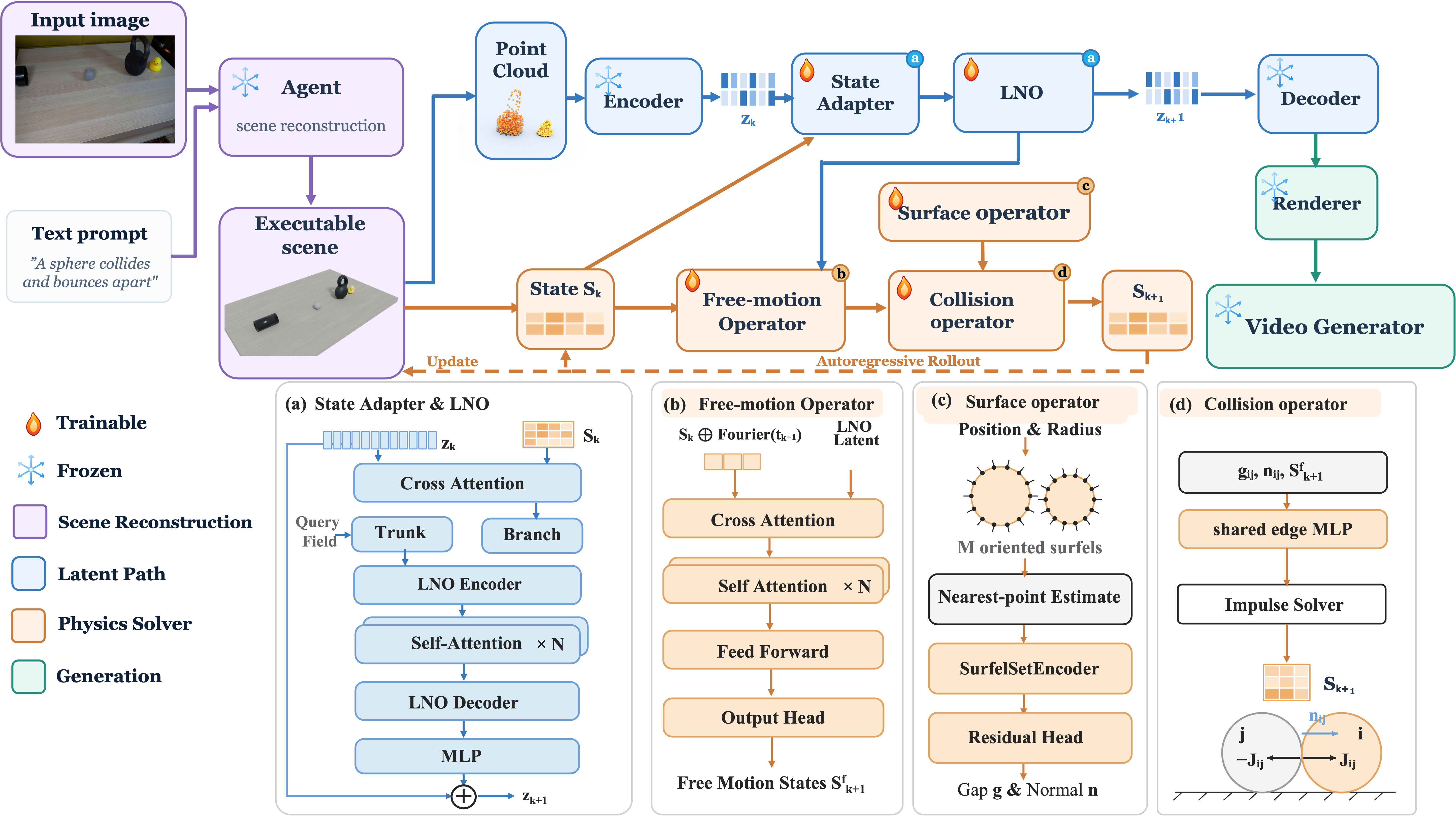}
  \caption{\textbf{Overview of \method{}.} 
  Stage~1 pretrains the geometry variational autoencoder and the surface operator.
  Stage~2 freezes them and trains the neural operator on rollouts of frames. At inference the operator rolls out the scene, and the decoded occupancy is rendered into a control video for the frozen Wan2.1-VACE~\citep{jiang2025vace}. 
  (a) State adapter and latent neural operator. 
  (b) Free-motion proposal. 
  (c) Surface operator.
  (d) Collision operator: learned contact probabilities and coefficients enter a closed-form impulse update.
  Snowflakes mark frozen modules, and flames mark trainable parameters.}
  \label{fig:pipeline}
\end{figure*}

\subsection{Approximation of Latent Neural Operators}
\label{sec:theoretical-boundary}

Inspired by the function-space analysis of latent nonlinear operators between Hilbert spaces~\citep{yang2026variationspacel}, we formulate the upper boundary on the approximation error of latent neural operator approximation. We first establish an structure-independent approximation error boundary, then specialize it to attention-based latent transitions, and finally discuss its implications for training.

We model the physical flow with a latent neural operator~\citep{wang2024latent},
\begin{equation}
    \mathcal{L}_N = \mathcal{D}\circ\mathcal{T}_N\circ\mathcal{E},
    \label{eq:lno-physical-factorization}
\end{equation}
where $\mathcal{E}:\mathcal{U}\rightarrow\mathcal{Z}$ encodes the observed history, $\mathcal{T}_N:\mathcal{Z}\rightarrow\mathcal{Z}$ evolves the physics in a learned latent space, and $\mathcal{D}:\mathcal{Z}\rightarrow\mathcal{V}$ decodes the predicted evolution. We assume $\mathcal{U}$ and $\mathcal{V}$ are separable real Hilbert spaces, which is usually plausible in physical scenes.
Let $\mathcal{S}^{\dagger}:\mathcal{U}\to\mathcal{V}$ denote ideal PDE-induced one-step flow map, and let
$\rho$ be the input distribution. Our objective is to characterize the physical approximation error.

\begin{definition}[Approximation Error]
    \label{def:latent-error}
    The physical approximation error is defined in Eq.~\eqref{eq:physical-approx-error}
    \begin{equation}
        \varepsilon_{\mathrm{phys}} =\left\| \mathcal{S}^{\dagger}-\mathcal{L}_N
        \right\|_{L_\rho^2(\mathcal{V})}
        \label{eq:physical-approx-error}
    \end{equation}
    Let $H:=L^2_\mu(\mathcal Z;\mathcal Z)$ and let $\mathcal T^\dagger\in H$ be the ideal latent transition. Denote $(\Omega,\mathfrak A)$ as a measurable space and $\mathcal F:\mathcal Z\times\mathcal Z\to\mathcal Z$ as a measurable update mapping. For a budget $M>0$, the \emph{representation error} is
    \begin{equation}
        \delta_M :=
        \inf_{\substack{\nu\in\mathcal M(\Omega)\\ \|\nu\|_{\rm TV}\leq M}}
        \Bigl\|
            \mathcal T^\dagger
            -\mathcal F\Bigl(\cdot,\int_\Omega\phi_\omega\,{\rm d}\nu(\omega)\Bigr)
        \Bigr\|_H
        \label{eq:latent_representation}
    \end{equation}
    where $\mathcal M(\Omega)$ denotes finite signed measures on
    $(\Omega,\mathfrak A)$ and the integral is a Bochner integral in $H$.
    For a model class $\mathfrak H_N\subset H$, the
    \emph{realization error} is
    \begin{equation}
        \begin{aligned}
            \eta_{N,M} &:= \sup_{g\in\Sigma_{N,M}}\ \inf_{\mathcal T\in\mathfrak H_N}
            \|\mathcal F(\cdot,g)-\mathcal T\|_H \\
            \Sigma_{N,M} &:= \Bigl\{\sum_{j=1}^{N}c_j\phi_{\omega_j}:\
            \omega_j\in\Omega,\ c_j\in\mathbb R,\ \sum_{j=1}^{N}|c_j|\leq M\Bigr\}
        \end{aligned}
         \label{eq:latent_realization}     
    \end{equation}
\end{definition}

Then we have the approximation boundary of Latent Neural Operators:
\begin{theorem}[Approximation Error Boundary]
    \label{thm:lno_bound}
    Let $\mathcal S^\dagger\in L^2_\rho(\mathcal V)$ and $\mu=\mathcal E_\#\rho$.
    Let $\mathcal T^\dagger$, $\mathfrak H_N$, $\mathcal F$ and the dictionary
    $\mathcal A$ be as in Definition~\ref{def:latent-error}, and let
    $\mathcal D$ be $L_{\mathcal D}$-Lipschitz. Then for every $N\geq1$ and
    $M>0$,
    \begin{equation}
        \begin{aligned}
            \inf_{\mathcal T_N\in\mathfrak H_N} \Bigl\|\mathcal S^\dagger &-\mathcal D\circ\mathcal T_N\circ\mathcal E\Bigr\|_{L_\rho^2}
            \leq \varepsilon_{\rm fac} \\
            & + L_{\mathcal D}\!\left(
                \delta_M + \frac{L_{\mathcal F}\Gamma_{\mathcal A}M}{\sqrt N} + \eta_{N,M}
            \right)
        \end{aligned}
        \label{eq:lno_general_bound}
    \end{equation}
    where
    $\varepsilon_{\rm fac}:=\|\mathcal S^\dagger-\mathcal D\circ\mathcal T^\dagger\circ\mathcal E\|_{L_\rho^2}$.
\end{theorem}
Theorem~\ref{thm:lno_bound} splits the physical approximation error into four parts: latent factorization ($\varepsilon_{\rm fac}$, set by the encoder-decoder pair), operator representation ($\delta_M$, set by the expressiveness of the dictionary), finite-width sampling ($N^{-1/2}$ term), and model realization ($\eta_{N,M}$, the mismatch between the architecture and the ideal $N$-term mixture). For specific attention-based neural operators~\citep{wang2024latent}, we prove in Appendix~\ref{app:latent-theory}.

Theorem~\ref{thm:lno_bound} motivates a two-stage training strategy by decoupling representation learning from latent dynamics modeling. In the first stage, the encoder and decoder are trained to construct a compact latent space that preserves essential physical information. In the second stage, the neural operator learns the underlying dynamics within this latent space, avoiding the need to jointly optimize representation and temporal evolution from scratch. Separating these objectives provides a practical way to balance physical information preservation and operator learnability. And geometry space is a natural choice due to its rich hidden states. \method{} adopts a two stage training pipeline, similar to some recent research~\citep{zhang2025cheapgeometryexpensivephysics,medranonavarro2026geometrymeetsphysicsdataefficient}.

\section{Framework}
\label{sec:method}

\subsection{Overview}
\label{sec:method-overview}

\begin{table*}[!tb]
    \centering\small
    \caption{\textbf{Physical plausibility on the subset of Physics-IQ-Verified~\citep{raedsch2026physicsiqverified} and RigidBench~\citep{jain2026rigidbench}.} Physics-IQ-Verified block uses the rigid-body subset. Both are tested on four random seeds (0, 42, 1989704766, 177674710). ``--'' means there is no compatible result.}
    \label{tab:main}
    \setlength{\tabcolsep}{3.5pt}
    \resizebox{\linewidth}{!}{
    \begin{tabular}{lc|ccc|cccccc}
        \toprule
        & \multicolumn{1}{c|}{ } & \multicolumn{3}{c|}{Physics-IQ-Verified subset} & \multicolumn{6}{c}{RigidBench~\citep{jain2026rigidbench}} \\
        Method &  \#Params & Score $\uparrow$ & ST-IoU $\uparrow$ & MSE $\downarrow$
        & IoU$\uparrow$& L2$\downarrow$ & Cham.$\downarrow$ & ATE$\downarrow$ & ATE-3D$\downarrow$ & SI-MSE$\downarrow$ \\
        \midrule
        CogVideoX-5B-I2V~\citep{yang2024cogvideox} & 5B & 30.59 & .297 & \best{.0046} & .146 & .222 & .361 & .186  & 2.63 & .0349   \\
        HunyuanVideo1.5-480P-I2V~\citep{wu2025hunyuanvideo15} & 8.3B & 30.08 & .151 & .0057  & .170 & .211 & .345 & .181 & 2.19 & .0372 \\
        Wan2.2-TI2V~\citep{wan2025} & 5B & 24.70 & .165 & .0057 & .183 & .220 & .370 & {.200} & 2.64 & \best{.0288} \\
        Cosmos-Predict-2.5-2B~\citep{nvidia2026cosmos} & 2B  & 27.05 & .190 & .0060  & .171 & .300 & .527 & .282 & 2.84 & .0331   \\
        \midrule
        PhysGen~\citep{liu2024physgen} & 0.9B & - & -  & - & .106 & .272 & .464 & .190 & 2.63 & .0338 \\
        PhyT2V~\citep{xue2025phyt2v} & 5B & 12.09 & .196 & .165 & .124 & .240 & .409 & .287 & 4.36 & .1631 \\        
        NewtonGen~\citep{yuan2026newtongen} & 5B & -  & - & - & .066 & .211 & .339 & .225 & 2.78 & .2028 \\
        VideoREPA~\citep{zhang2025videorepa} & 5B & 15.64  & .313 & .129 & .117 & .264 & .451 & .279 & 3.42 & .1745 \\
        \midrule
        Wan2.1-VACE~\citep{jiang2025vace} & 1.3B & 20.05 & .178 & .0112 & .167 & .229 & .382 & .180 & 2.71 & .0323 \\
        Wan2.1-VACE~\citep{jiang2025vace} \textbf{+ \method{}} & 1.3B & 30.27\up{10.22} & .270\up{.092} & .0095\up{.0017} & .272\up{.105} & \best{.148\up{.081}} & \best{.238\up{.144}} & \best{.152\up{.028}} & 2.77\down{.06} & .0307\up{.0016} \\
        MiniMax-H3~\citep{minimax2026h3} & 33B & 39.36 & .236 & .0048 & .202 & .198 & .328 & .186 & \best{2.00} & .0375 \\
        MiniMax-H3~\citep{minimax2026h3} \textbf{ + \method{}} & 33B & \best{44.62\up{5.26}} & \best{.424\up{0.188}} & {.0051\down{.0003}} & \best{.303\up{.101}} & .150\up{.048} & .241\up{.087} & .174\up{.012} & 2.01\down{.01} & .0311\up{.064}\\
        \bottomrule
    \end{tabular}}
\end{table*}

\method{} separates physical evolution from video synthesis (Fig.~\ref{fig:pipeline}). Given a text description $\mathbf T$ and an input image $\mathbf I_{ref}$, an Astra~\citep{openai2026gpt6astra}-assisted executable scene reconstruction agent estimates object identities, initial geometry, camera and physical attributes. The dynamics model evolves a latent scene representation together with explicit object states. Its predicted positions drive a depth sequence, which conditions a pretrained video model. The same dynamics can therefore be paired with different video backends.

For each dynamic object, the operator uses $\mathbf S_i=(\mathbf p_i,\mathbf v_i,\boldsymbol\omega_i,r_i,a_i, \mathbf u_i)\in\mathbb R^{14}$,
where $r_i,a_i$ are axisymmetric proxy radii and $\mathbf u_i$ is a unit symmetry axis. Mass and material coefficients are separate attributes. The evaluated sphere model replaces the proxy by its volume-equivalent sphere, $r_i=a_i=(r_i^2a_i)^{1/3}$, with a fixed axis convention. Reconstruction scale, unobserved velocity components, and material values supplied by a prior are recorded as assumptions rather than image-measured quantities.

\textbf{Scene representation.} A frozen occupancy encoder~\citep{zhang2023shape2vecset} maps 2048
scene-level proxy-volume samples to $\mathbf Z\in\mathbb R^{256\times32}$.
Each input sample has four coordinates $(x,y,z,t)$, with $t=0$ at initialization. These samples describe the dynamics proxies; the original appearance meshes are retained separately for rendering. The occupancy encoder is trained on ShapeNetCorev2~\citep{shapenet2015}.

\textbf{Free-Motion Proposal.}
At step $k$, a state-to-latent attention module writes the current object states into the scene slots. A latent neural operator~\citep{wang2024latent} then updates the scene representation:
\begin{equation}
    \mathbf Z_{k+1}=\mathbf Z_k+
    \mathcal T_\theta\!\left(\boldsymbol\xi,
    \operatorname{LN}\big(\mathcal A_\theta(\mathbf Z_k,\mathbf S_k)\big)\right).
    \label{eq:state-update}
\end{equation}
Here $\boldsymbol\xi$ is a fixed slot coordinate as the query field, not a physical location. $T_\theta$ is the transformer-based latent neural operator~\citep{wu2023latent}. Object queries attend to the evolved slots and to other objects. A bounded head predicts nine dimensionless residuals
$(\boldsymbol\delta_p,\boldsymbol\delta_v,\boldsymbol\delta_\omega)$ per object. With $\Delta t=1/60$\,s and training-set increment scales $\boldsymbol\sigma_v,\boldsymbol\sigma_\omega$, the proposal $\mathcal{P}_{\mathrm{free}}$ is
\begin{equation}
    \begin{aligned}
    \widetilde{\mathbf p}&=\mathbf p+\Delta t\,\mathbf v
       +\tfrac12\Delta t^2\mathbf g
       +\Delta t(\boldsymbol\sigma_v\odot\boldsymbol\delta_p),\\
    \widetilde{\mathbf v}&=\mathbf v+\Delta t\,\mathbf g
       +\boldsymbol\sigma_v\odot\boldsymbol\delta_v,\\
    \widetilde{\boldsymbol\omega}&=\boldsymbol\omega
       +\boldsymbol\sigma_\omega\odot\boldsymbol\delta_\omega.
    \end{aligned}
    \label{eq:m32-proposal}
\end{equation}
Each later step consumes the previous predicted state and latent. Future reference positions or occupancy codes are supervision targets, never rollout inputs.

\textbf{Contact response.}
The free-motion proposal is followed by static-surface response, dynamic-pair response, and a positional support correction. Finite geometric queries determine gaps, normals, and contact candidates. Learned coefficient networks adjust friction and restitution within fixed bounds; analytic impulse updates use the proxy mass and inertia. Static and pair response use 24- and 26-dimensional feature vectors, respectively. A finite-patch environment and analytic proxy contacts provide the physical scaffold; the model does not perform mesh-aware collision detection on the appearance meshes.

\textbf{Physics-Video Interface.}
We retain the reconstructed mesh at the initial time and bind each dynamic component to its predicted object centre. Its vertices follow
\begin{equation}
    \mathbf x_i(t)=\mathbf x_i(0)+\widehat{\mathbf p}_i(t)-\widehat{\mathbf p}_i(0).
    \label{eq:mesh-binding}
\end{equation}
Static geometry remains fixed. This interface applies centre translation; it does not apply predicted mesh rotations. camera-space triangle rasterization produces positive camera-$z$ depth, instance ownership, and no-hit maps. A common inverse-depth scale over the clip yields a three-channel grayscale condition, with no-hit pixels black. Thus the video condition is rendered from predicted geometry, rather than obtained by decoding the occupancy latent into an image. For Wan2.1-VACE-1.3B~\citep{jiang2025vace}, the condition contains real RGB input at initial time and predictes depth thereafter. The native mask is zero at the first frame and one for future frames. We also use SolarRenderer with its native MiniMax-H3 backbone~\citep{minimax2026h3}. It receives the same initial RGB and predicted depth through its native encoding pipeline. The video backends are frozen for these evaluations.

\subsection{Training}
\label{sec:method-training}

\begin{figure*}[t]
    \centering
    \includegraphics[width=\textwidth,height=0.85\textheight]{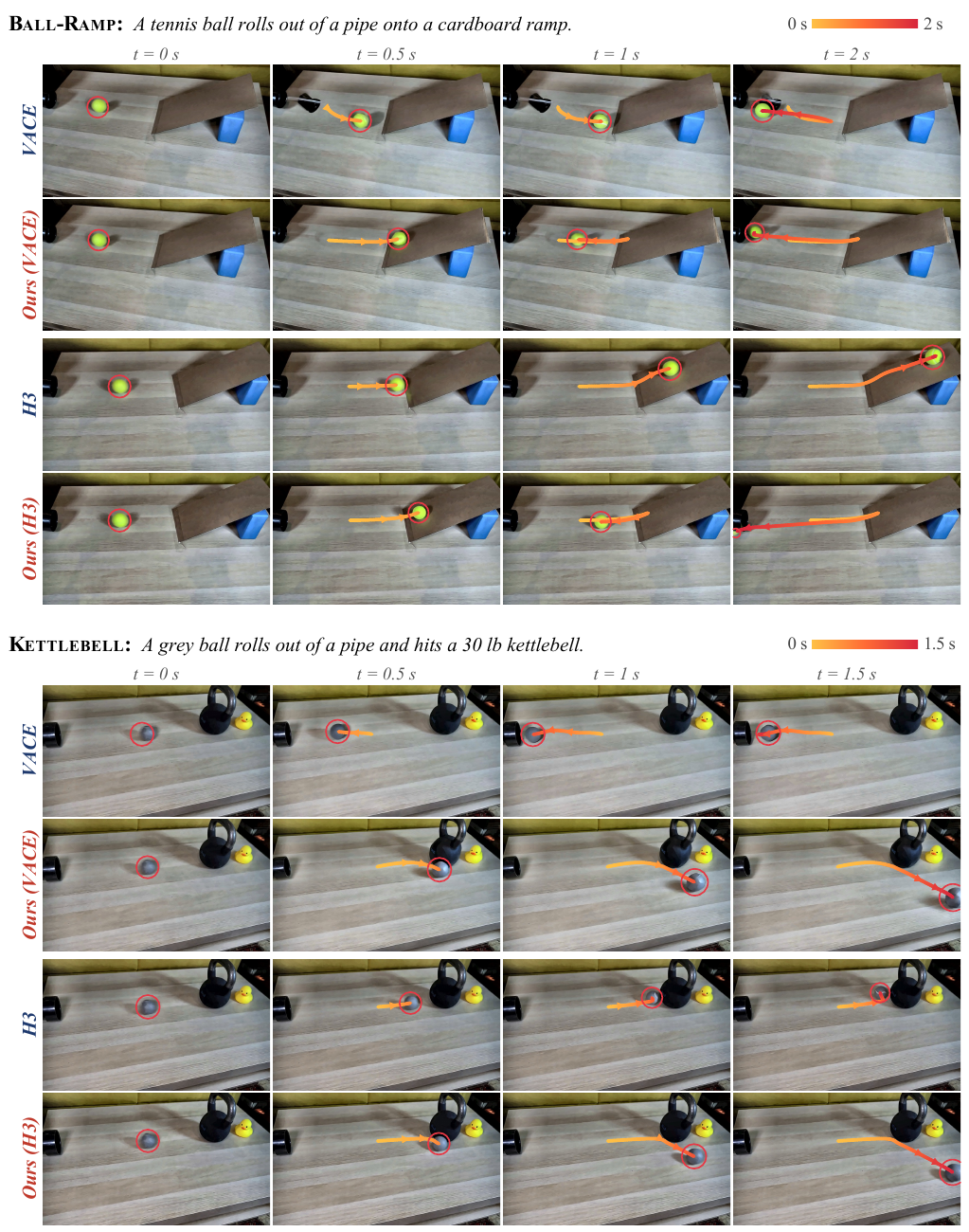}
    \caption{\textbf{Qualitative comparison on two Physics-IQ-Rigid scenes.}
    Each block pairs a native video backbone with the same backbone driven by \method{}: VACE-1.3B vs.\ Ours (VACE), and MiniMax-H3 vs.\ Ours (H3, with SolarRenderer). The baselines receive the first frame and the text prompt; \method{} additionally uses the reconstructed initial state and its predicted depth condition. Overlays show the ball trajectory from $t=0$ up to the displayed time, tracked in each generated video with SAM~2~\citep{ravi2024sam2}; color encodes time and the red circle marks the current position. In \textsc{Ball-Ramp}, the VACE rows use the right camera view and the H3 rows the center view of the same scenario.}
    \label{fig:bench-qualitative}
\end{figure*}

The modules are trained in stage-wise pattern, each frozen before the next, and the operator is never trained through the video objective: with the physical loss active the diffusion gradient leaves it unchanged, and without it the operator drifts toward more motion and nearly doubles its short-horizon error. 

\textbf{Stage~1} pretrains the surface operator on exact gap and normal targets, and the occupancy autoencoder on space-time segments with the binary cross-entropy loss~\citep{mescheder2019occupancy}. Details are shown in Appendix~\ref{app:impl-details}

\textbf{Stage~2} freezes both and trains $\mathcal{A}_{\mathrm{c}}$, free-motion operator $\mathcal{P}_{\mathrm{free}}$, and collision operator $\mathcal{C}_{\mathrm{col}}$ jointly on free-running rollouts of $K$ frames from a single initial frame, with
\begin{equation}
    \mathcal L=\mathcal L_p+\mathcal L_v+\mathcal L_\omega
       +0.2\mathcal L_u +0.1\mathcal L_z+0.1\mathcal L_{\mathrm{occ}}
    \label{eq:latent-dynamics-loss}
\end{equation}

Position, velocity, angular velocity, and latent errors use normalized smooth-$L_1$ losses. Axis error is sign-invariant and masked when the axis is unobservable; occupancy uses binary cross-entropy on query points. All reductions respect valid objects and times. Increment normalization uses training data only. We select the sphere checkpoint at update 13000 using the prescribed full-validation trajectory criterion and freeze it for video generation. This method uses pretrained video conditioning; it does not require a new video-adapter training stage.

\section{Experiments}
\label{sec:experiments}

\subsection{Setup}
\label{sec:exp-setup}

\textbf{Data and Preprocessing} 
We bulid about $6,000$ rigid-body videos via the data generation pipeline Kubric~\citep{greff2022kubric}. Our neural operator is trained on rigid-body trajectories 
from our own data. And we train the 3DPoint variational encoder and decoder from MOVi-Spheres~\citep{HOPNet2025} and ShapeNetCore~\citep{shapenet2015}. Appendix~\ref{app:impl-details} lists more details. For evaluation, we utilize two rigid-body-focused video benchmarks: a subset of Physics-IQ-Verified~\citep{raedsch2026physicsiqverified} a general benchmark of physical reasoning, from which we select the scenes whose prompts describe rigid-body motion such as collision, rolling, sliding, and free fall, and finally get 81 five-second videos, annotated as \textit{ Physics-IQ-Rigid}; and RigidBench~\citep{jain2026rigidbench}, a public benchmark built for rigid-body dynamics, which provides 100 two-second videos. Our scene reconstruction uses an Astra~\citep{openai2026gpt6astra}-assisted executable scene program. Details are shown in Appendix~\ref{app:impl-details}. 

\textbf{Metrics.}
We choose serval critical metrics which evaluating the physical plausibility of rigid-body motion in the video from Physics-IQ~\citep{motamed2026physics,raedsch2026physicsiqverified} and RigidBench~\citep{jain2026rigidbench}. From Physics-IQ, we select Physics-IQ score (maximum:100) representing overall performance; SpatioTemporal-IoU and MSE measuring accuracy of spatiotemporal motion trajectories. In RigidBench~\citep{jain2026rigidbench}, we report per-object mask IoU, the mask-centroid error (L2) and the bidirectional Chamfer distance, both normalized by image height~\citep{li2025pisa}. We also report the 2D and 3D trajectory error of tracked points (ATE \& ATE-3D), and the scale-invariant depth error (SI-MSE). We follow the definitions and implementations of the respective benchmark protocols.

\textbf{Baselines.}
First, we compare \method{} with physics-aware video generators cover the three places where physics is injected: language reasoning~\citep{xue2025phyt2v}; explicit states~\citep{liu2024physgen,yuan2026newtongen}; and representation alignment~\citep{zhang2025videorepa}.  Then we also choose general video generators of comparable scale, to complete the physical video quality. These frozen backbone without a physical condition isolates what the neural operators adds. 

\textbf{Implementation.}
In \method{}, we choose Wan2.1-VACE-1.3B~\citep{wan2025,jiang2025vace} with frozen weights as the video generator. Sampling uses 30 steps at $832\times480$ with classifier-free guidance $5.0$ on the text and none on the physical condition; clips have 81 frames at 16\,fps on Physics-IQ and 21 frames on MOVi-Spheres. The simulator is frozen after Stage~2, advances one block per step, and is decoded at every video timestamp. All experiments are conducted on 8 Nvidia H100 GPUs. The implement details of \method{} and other baselines are shown in Appendix~\ref{app:impl-details}.

\subsection{Main Results}
\label{sec:exp-main}

\paragraph{Qualitative comparison.}
\label{sec:exp-qualitative}

Figure~\ref{fig:bench-qualitative} pairs each backbone with its \method{}-driven counterpart on two Physics-IQ-Rigid scenes. In \textsc{Ball-Ramp}, the native VACE ball stays near the pipe and never climbs the ramp, and the native H3 ball keeps climbing to the top of the ramp within 2\,s. With \method{}, both backbones show the ball climbing the ramp, decelerating, and rolling back toward the pipe. In \textsc{Kettlebell}, native VACE rolls the ball back into the pipe and native H3 stops it at the kettlebell, whereas with \method{} the ball is deflected on contact and leaves the frame. The baselines use only the first frame and text, and the examples were selected for visualization; they illustrate the effect of the dynamics condition rather than establish benchmark-wide rankings.

\subsection{Ablations}
\label{sec:exp-ablation}

\begin{table}[t]
    \centering\small
    \caption{\textbf{Shared latent-slot early-optimization pilot.}
    All learned variants use 50 updates and seed 0; rollouts span 20 steps at
    60\,Hz. Test errors use the same 16 scenes, normalized by equivalent radius
    $R_{\mathrm{eq}}$ (floor 1\,mm). Parameters count the active processor.}
    \label{tab:ablation-q}
    \setlength{\tabcolsep}{3pt}
    \begin{tabular}{lrrr}
    \toprule
    Dynamics & Params (M) & ADE/$R_{\mathrm{eq}}\downarrow$ & FDE/$R_{\mathrm{eq}}\downarrow$\\
    \midrule
    Ours & \textbf{1.283}& \textbf{0.5181} & \textbf{0.9699}\\
    MPM~\citep{blatny2025mpm}  & 0 & 0.5204 & 0.9823\\
    Transolver~\citep{wu2024transolver} & 1.298 & 0.5189 & 0.9715\\
    GNOT-CGPT~\citep{hao2023gnot} & 1.338 & 0.5184 & 0.9702\\
    \bottomrule
    \end{tabular}
\end{table}

The small-scale study in Table~\ref{tab:ablation-q} separates the learned dynamics from the
analytic prior and compares LNO with MPM~\citep{blatny2025mpm}, Transolver~\citep{wu2024transolver} and GNOT-CGPT~\citep{hao2023gnot} processors adapted to a shared latent-slot interface.
Only the processor is replaced for the learned variants. The pretrained occupancy encoder is shared and kept frozen; the state-head architecture, integrator, and contact rules are also shared. The analytic variant retains the same physical priors with all learned corrections zeroed. All three learned models start from matched shared-head weights and use the same 50 windows from a 32-scene training pool (23 actually observed), with 8 validation and 16 test scenes. They use dataset initial states, static geometry, state-only loss, and causal rollouts; no production checkpoint is substituted for the pilot Full.

All four variants achieve similar prediction errors under the same training budget. The full model yields slightly lower test ADE and FDE than the alternatives, although their rollout error curves largely overlap. These preliminary results, obtained with a single random seed and only 50 training updates, do not establish convergence or statistically significant differences. They also do not support conclusions about the relative performance of the final models or improvements in downstream video control. This state-space evaluation is distinct from the Physics-IQ evaluation and the partially completed RigidBench evaluation.

\section{Conclusion}
\label{sec:conclusion}

\method{} separates rigid-body state prediction from visual synthesis. A latent neural operator produces object-state rollouts, and frozen video diffusion models generates aligned videos. We evaluate video agreement on the strict 81 Physics-IQ subset and RigidBench. These results shows the current pipeline is a promising research direction . However, state-estimation uncertainty, contact and support failures, and imperfect object persistence remain limitations; stable long-horizon physical control is an open goal.
\section*{AI Use Statement}

We used generative AI tools to assist with English-language polishing,
code development and debugging, and discussions related to theoretical
formulations and derivations. AI-generated suggestions were used only as
research assistance; all mathematical claims, derivations, code, and
experimental implementations were independently checked and verified by
the authors. The authors take full responsibility for the final content
of this work.
{\small
\bibliographystyle{ieeenat_fullname}
\bibliography{iclr2027_conference}
}
\clearpage
\appendix
\section{More Proof}
\label{app:latent-theory}

\subsection{Definition and Setting}
\label{app:theory-setting}

Throughout, $\mathcal Z$ is a separable real Hilbert space, $\mu=\mathcal E_{\#}\rho$, and
\begin{equation}
    \mathcal X := L^2_\mu(\mathcal Z;\mathcal Z), \qquad
    \|f\|_{\mathcal X}^2 := \int_{\mathcal Z}\|f(z)\|_{\mathcal Z}^2\,{\rm d}\mu(z).
\end{equation}
Since $\mathcal Z$ is a separable Hilbert space, so is $\mathcal X$. We use the objects of Definition~\ref{def:latent-error} under the following standing assumptions.
\begin{itemize}
    \item \emph{Dictionary.} $(\Omega,\mathfrak A)$ is a measurable space, $\omega\mapsto\phi_\omega\in\mathcal X$ is Bochner measurable, and
    \begin{equation}
        \Gamma_{\mathcal A}:=\sup_{\omega\in\Omega}\|\phi_\omega\|_{\mathcal X}<\infty.
        \label{eq:app_dictionary_bound}
    \end{equation}
    For a finite signed measure $\nu$ on $\Omega$, the Bochner integral
    \begin{equation}
        g_\nu := \int_\Omega\phi_\omega\,{\rm d}\nu(\omega)\in\mathcal X
        \label{eq:app_integral_representation}
    \end{equation}
    is well defined, and $\|g_\nu\|_{\mathcal X}\leq\Gamma_{\mathcal A}\|\nu\|_{\rm TV}$.
    \item \emph{Update map.} $\mathcal F$ is measurable, $\mathcal F(\cdot,0)\in\mathcal X$, and $\mathcal F$ is $L_{\mathcal F}$-Lipschitz in its second argument, uniformly in the first.
    \item \emph{Encoder and decoder.} $\mathcal E:\mathcal U\to\mathcal Z$ is measurable, $\mathcal S^\dagger\in L^2_\rho(\mathcal V)$, $\mathcal T^\dagger\in\mathcal X$, and $\mathcal D$ is $L_{\mathcal D}$-Lipschitz.
\end{itemize}
With $\nu=0$ in \eqref{eq:latent_representation}, $\delta_M\leq\|\mathcal T^\dagger-\mathcal F(\cdot,0)\|_{\mathcal X}<\infty$ for every $M>0$.

\begin{lemma}[Basic Facts]
\label{lem:elementary}
\leavevmode
    \begin{enumerate}
        \item[(i)] For $g,g'\in\mathcal X$, we have $\mathcal F(\cdot,g)\in\mathcal X$ and
        $\|\mathcal F(\cdot,g)-\mathcal F(\cdot,g')\|_{\mathcal X}\leq L_{\mathcal F}\|g-g'\|_{\mathcal X}$.
        \item[(ii)] For $f_1,f_2\in\mathcal X$, the compositions $f_i\circ\mathcal E$ are well defined $\rho$-a.e.\ and
        \begin{equation}
            \|\mathcal D\circ f_1\circ\mathcal E-\mathcal D\circ f_2\circ\mathcal E\|_{L^2_\rho(\mathcal V)}
            \leq L_{\mathcal D}\|f_1-f_2\|_{\mathcal X}.
        \end{equation}
    \end{enumerate}
\end{lemma}

\begin{proof}
\leavevmode\par
(i) Pointwise, $\|\mathcal F(z,g(z))\|\leq\|\mathcal F(z,0)\|+L_{\mathcal F}\|g(z)\|$ and $\|\mathcal F(z,g(z))-\mathcal F(z,g'(z))\|\leq L_{\mathcal F}\|g(z)-g'(z)\|$. Square and integrate against $\mu$.

(ii) A $\mu$-null set has $\rho$-null preimage under $\mathcal E$ because $\mu=\mathcal E_{\#}\rho$, so $f_i\circ\mathcal E$ does not depend on the representative of $f_i$. By the Lipschitz property of $\mathcal D$ and the change of variables $z=\mathcal E(u)$,
\begin{equation}
    \begin{aligned}
    \int_{\mathcal U}\|\mathcal D(f_1(\mathcal E(u)))-\mathcal D(f_2(\mathcal E(u)))\|^2\,{\rm d}\rho(u) \\
    \leq L_{\mathcal D}^2\int_{\mathcal Z}\|f_1(z)-f_2(z)\|^2\,{\rm d}\mu(z)
    \end{aligned}
\end{equation}

\end{proof}
 
The following lemma is the source of the $N^{-1/2}$ rate.
\begin{lemma}[Finite-atom approximation]
\label{lem:finite_atom}
Assume \eqref{eq:app_dictionary_bound}. For every finite signed measure $\nu$ with $\|\nu\|_{\rm TV}\leq M$ and every $N\geq1$, there exists $g_N=\sum_{j=1}^N c_j\phi_{\omega_j}\in\Sigma_{N,M}$ such that
\begin{equation}
    \|g_\nu-g_N\|_{\mathcal X}\leq\frac{\Gamma_{\mathcal A}M}{\sqrt N}.
    \label{eq:app_finite_atom_bound}
\end{equation}
\end{lemma}

\begin{proof}
Let $a:=\|\nu\|_{\rm TV}$. If $a=0$, take $g_N=0$. Otherwise, the polar decomposition gives ${\rm d}\nu=a\,s\,{\rm d}\pi$, where $\pi:=|\nu|/a$ is a probability measure on $\Omega$ and $|s|=1$ $\pi$-a.e. Hence $g_\nu=\mathbb E_\pi[Y]$ with
\[
    Y(\omega):=a\,s(\omega)\,\phi_\omega, \quad \|Y(\omega)\|_{\mathcal X}\leq a\Gamma_{\mathcal A}
\]
Draw $\omega_1,\dots,\omega_N$ i.i.d.\ from $\pi$, put $Y_j:=Y(\omega_j)$ and
\begin{equation}
    g_N:=\frac1N\sum_{j=1}^N Y_j=\sum_{j=1}^N c_j\phi_{\omega_j}, \quad c_j:=\frac{a\,s(\omega_j)}{N}
    \label{eq:app_sampled_operator}
\end{equation}
Then $\sum_j|c_j|=a\leq M$, so $g_N\in\Sigma_{N,M}$, and $\mathbb E\,g_N=g_\nu$. In the Hilbert space $\mathcal X$ the centered variables $Y_j-g_\nu$ are independent with mean zero, so the cross terms vanish and
\begin{equation}
    \begin{aligned}
        \mathbb E\|g_N-g_\nu\|_{\mathcal X}^2
        =\frac1N\bigl(\mathbb E\|Y\|_{\mathcal X}^2-\|g_\nu\|_{\mathcal X}^2\bigr)
        \leq\frac{M^2\Gamma_{\mathcal A}^2}{N}
    \end{aligned}
     \label{eq:app_sampling_variance}
\end{equation}
Since the expectation obeys this bound, at least one realization satisfies it. Taking square roots gives \eqref{eq:app_finite_atom_bound}.
\end{proof}

\subsection{Proofs of Theorem~\ref{thm:lno_bound}}
\label{app:proof_lno_bound}

\begin{proof}
For any $\mathcal T_N\in\mathfrak H_N$, the triangle inequality and Lemma~\ref{lem:elementary}(ii) give
\begin{equation}
    \|\mathcal S^\dagger-\mathcal D\circ\mathcal T_N\circ\mathcal E\|_{L^2_\rho}
    \leq\varepsilon_{\rm fac}+L_{\mathcal D}\|\mathcal T^\dagger-\mathcal T_N\|_{\mathcal X}
\end{equation}
Taking the infimum over $\mathcal T_N\in\mathfrak H_N$ and proves \eqref{eq:lno_general_bound}.
\end{proof}
 
Theorem~\ref{thm:lno_bound} bounds the best achievable error in $\mathfrak H_N$, not the error of a trained model. Neither $\mathcal E$ nor $\mathcal D$ needs to be invertible; information lost by the encoder that prevents an accurate latent factorization is accounted for by $\varepsilon_{\rm fac}$. The bound has no explicit factor of $\dim(\mathcal Z)$, but dimension may enter implicitly through $\Gamma_{\mathcal A}$, $M$, $L_{\mathcal F}$, $L_{\mathcal D}$ and the representation errors.

\subsection{Error Boundary of Transformer-Specific Neural Operators}
We now specialize to attention-based latent neural operators. Let $J:\mathcal Z\to\mathbb R^{m\times d}$ be a tokenization map and define the attention head
\begin{equation}
    \begin{aligned}
        H_\omega(z) &= R_\omega\bigl(A_\omega(X)\,XW^V_\omega\bigr), \qquad X=J(z),\\
        A_\omega(X) &= \operatorname{softmax} \Bigl(\tfrac{(XW^Q_\omega)(XW^K_\omega)^\top}{\sqrt{d_k}}+B_\omega\Bigr),
    \end{aligned}
    \label{eq:latent_attention}
    \end{equation}
    where $W^V_\omega\in\mathbb R^{d\times d_v}$ and $R_\omega:(\mathbb R^{m\times d_v},\|\cdot\|_F)\to\mathcal Z$ is linear.

\begin{lemma}[Uniform boundary of softmax attention]
\label{lem:attention_boundedness}
Let $A\in\mathbb R^{m\times m}$ be a row-stochastic matrix, i.e.\ $a_{ij}\geq0$ and $\sum_j a_{ij}=1$. Then for every $V\in\mathbb R^{m\times d_v}$,
\begin{equation}
    \|AV\|_F\leq\sqrt m\,\|V\|_F.
    \label{eq:app_softmax_bound}
\end{equation}
\end{lemma}
 
\begin{proof}
Let $v_j$ be the $j$-th row of $V$, so that $(AV)_{i,:}=\sum_j a_{ij}v_j$ is a convex combination of the rows. By convexity of $\|\cdot\|_2^2$ (Jensen),
\[
    \Bigl\|\sum_{j=1}^m a_{ij}v_j\Bigr\|_2^2\leq\sum_{j=1}^m a_{ij}\|v_j\|_2^2.
\]
Summing over $i$ and using $\sum_i a_{ij}\leq m$ (as $0\leq a_{ij}\leq1$),
\[
    \|AV\|_F^2\leq\sum_{j=1}^m\Bigl(\sum_{i=1}^m a_{ij}\Bigr)\|v_j\|_2^2\leq m\|V\|_F^2. \qedhere
\]
\end{proof}

\begin{corollary}[Attention-based latent neural operator boundary]
    \label{cor:attention_bound}
    Assume $\mu$ has a finite second moment, $\|J(z)\|_F\leq B_J(1+\|z\|_{\mathcal Z}^2)^{1/2}$,
    $\omega\mapsto H_\omega\in L^2_\mu(\mathcal Z;\mathcal Z)$ is Bochner
    measurable, $\|W^V_\omega\|_{\rm op}\leq B_V$ and $\|R_\omega\|_{\rm op}\leq B_R$.
    Take the dictionary $\phi_\omega=H_\omega$ and the model class
    \begin{equation}
        \begin{aligned}
            \mathfrak H_N=\Bigl\{
                z\mapsto\mathcal F\Bigl(z,\textstyle\sum_{j=1}^N c_jH_{\omega_j}(z)\Bigr)
                :\ \omega_j\in\Omega, \\ \ c_j\in\mathbb R,\ \textstyle\sum_{j=1}^N|c_j|\leq M
            \Bigr\}
        \end{aligned}
    \end{equation}
    so that $\eta_{N,M}=0$. Then, under the assumptions of Theorem~\ref{thm:lno_bound},
    \begin{equation}
        \inf_{\mathcal T_N\in\mathfrak H_N}
        \left\|\mathcal S^\dagger-\mathcal D\circ\mathcal T_N\circ\mathcal E\right\|_{L_\rho^2}
        \leq
        \varepsilon_{\rm fac}
        + L_{\mathcal D}\,\delta_M^{\rm att}
        + \frac{L_{\mathcal D}L_{\mathcal F}M\,\Gamma_{\rm att}}{\sqrt N}
    \label{eq:attention_approx_bound}
    \end{equation}
    where $\delta_M^{\rm att}$ is \eqref{eq:latent_representation} with
    $\phi_\omega=H_\omega$ and
    \[
        \Gamma_{\rm att}
        :=\sqrt m\,B_RB_VB_J
        \Bigl(\int_{\mathcal Z}(1+\|z\|_{\mathcal Z}^2)\,{\rm d}\mu(z)\Bigr)^{1/2}.
    \]
\end{corollary}

We show that the attention-head dictionary in \eqref{eq:latent_attention} satisfies \eqref{eq:app_dictionary_bound}.
 
\begin{proof}
Here $R_\omega$ is a bounded linear map from $(\mathbb R^{m\times d_v},\|\cdot\|_F)$ to $\mathcal Z$ with operator norm at most $B_R$. Using Lemma~\ref{lem:attention_boundedness} with $V=J(z)W^V_\omega$, and $\|J(z)W^V_\omega\|_F\leq\|J(z)\|_F\|W^V_\omega\|_{\rm op}$,
\begin{equation}
    \|H_\omega(z)\|_{\mathcal Z}
    \leq\sqrt m\,B_RB_VB_J(1+\|z\|_{\mathcal Z}^2)^{1/2}.
    \label{eq:app_head_pointwise}
\end{equation}
Squaring and integrating against $\mu$ gives $\sup_\omega\|H_\omega\|_{\mathcal X}\leq\Gamma_{\rm att}$, which is finite because $\mu$ has a finite second moment. Hence (A1) holds for $\mathcal A_{\rm att}=\{H_\omega:\omega\in\Omega\}$ with $\Gamma_{\mathcal A}=\Gamma_{\rm att}$.
 
Every element of $\{\mathcal F(\cdot,g):g\in\Sigma_{N,M}\}$ belongs to the model class $\mathfrak H_N$ of the corollary by construction, so $\eta_{N,M}^{\rm att}=0$. Substituting $\Gamma_{\mathcal A}=\Gamma_{\rm att}$ and $\eta_{N,M}=0$ into Theorem~\ref{thm:lno_bound} gives \eqref{eq:attention_approx_bound}.
\end{proof}
 
No bound on $W^Q_\omega$, $W^K_\omega$ or $B_\omega$ is needed for the magnitude of $H_\omega$. The factor $\sqrt m$ is the sharp operator-norm boundary for row-stochastic matrices (attained when all rows put their mass on one token). Query and key projections still matter for the Lipschitz stability of the full transition, and for $\delta_M^{\rm att}$. The corollary concerns a fixed, uniformly bounded head dictionary; it does not give the same rate for arbitrary deep Transformers or for head dimensions that change with $N$.

\subsection{Rollout Error Boundary}
\label{app:rollout_training_proof}

Let $\{x_t\}_{t=0}^T\subset\mathcal V$ be a ground-truth trajectory, $z_t=\mathcal E(h_t)$ the encoding of the history $h_t$ ending at time $t$, $\hat z_0=z_0$, $\hat z_{t+1}=\mathcal T_N(\hat z_t)$ and $\hat x_t=\mathcal D(\hat z_t)$. The residuals $a_t,r_t$ are as in \eqref{eq:latent_residuals}. We assume that $\mathcal D$ is $L_{\mathcal D}$-Lipschitz and that, for some $L_T\geq0$,
\begin{equation}
    \|\mathcal T_N(z_t)-\mathcal T_N(\hat z_t)\|_{\mathcal Z}\leq L_T\|z_t-\hat z_t\|_{\mathcal Z},
    \quad t=0,\dots,T-1.
    \label{eq:app_local_stability}
\end{equation}
This holds in particular if $\mathcal T_N$ is $L_T$-Lipschitz on a set containing all $z_t$ and $\hat z_t$. We use the convention $0^0=1$.

Define
\begin{equation}
    a_t:=\|x_t-\mathcal D(z_t)\|_{\mathcal V}, \quad
    r_t:=\|z_{t+1}-\mathcal T_N(z_t)\|_{\mathcal Z}.
    \label{eq:latent_residuals}
\end{equation}
Then for any weights $w_t\geq0$,
\begin{equation}
    \sum_{t=1}^{T}w_t\|x_t-\hat x_t\|_{\mathcal V}^2
    \leq 2\sum_{t=1}^{T}w_ta_t^2 + 2L_{\mathcal D}^2\sum_{k=0}^{T-1}\beta_kr_k^2,\label{eq:rollout_training_bound}
\end{equation}
\begin{equation}
    \beta_k=\sum_{t=k+1}^{T}w_t\,L_T^{\,t-1-k}\sum_{j=0}^{t-1}L_T^{\,j}.
    \label{eq:propagation_weight}
\end{equation}

The weights $\beta_k$ grow geometrically when $L_T>1$, so the bound is
informative only for approximately non-expansive latent dynamics. It
motivates jointly minimizing reconstruction and latent-transition
residuals with horizon-dependent weights. Note that it controls pathwise
residuals, complementing the $L^2_\rho$ operator-level guarantee of
Theorem~\ref{thm:lno_bound}.

\begin{lemma}[Latent error propagation]
\label{lem:latent_rollout}
For every $1\leq t\leq T$,
\begin{equation}
    \|z_t-\hat z_t\|_{\mathcal Z}\leq\sum_{k=0}^{t-1}L_T^{\,t-1-k}r_k.
    \label{eq:app_latent_rollout}
\end{equation}
\end{lemma}

\begin{proof}
Let $e_t:=\|z_t-\hat z_t\|_{\mathcal Z}$, so $e_0=0$. Since $\hat z_{t+1}=\mathcal T_N(\hat z_t)$, the triangle inequality and \eqref{eq:app_local_stability} give
\[
    e_{t+1}\leq\|z_{t+1}-\mathcal T_N(z_t)\|+\|\mathcal T_N(z_t)-\mathcal T_N(\hat z_t)\|\leq r_t+L_Te_t.
\]
Induction on $t$ yields \eqref{eq:app_latent_rollout}.
\end{proof}

\begin{proof}[Proof]
By the triangle inequality, Lipschitz continuity of $\mathcal D$ and Lemma~\ref{lem:latent_rollout},
\[
    \|x_t-\hat x_t\|_{\mathcal V}\leq a_t+L_{\mathcal D}\|z_t-\hat z_t\|_{\mathcal Z}
    \leq a_t+L_{\mathcal D}\sum_{k=0}^{t-1}L_T^{\,t-1-k}r_k.
\]
Using $(a+b)^2\leq2a^2+2b^2$, then the Cauchy--Schwarz inequality $\bigl(\sum_k b_kr_k\bigr)^2\leq\bigl(\sum_k b_k\bigr)\bigl(\sum_k b_kr_k^2\bigr)$ with $b_k=L_T^{t-1-k}$ and $S_t:=\sum_{j=0}^{t-1}L_T^{\,j}$,
\begin{equation}
    \|x_t-\hat x_t\|_{\mathcal V}^2
    \leq2a_t^2+2L_{\mathcal D}^2\,S_t\sum_{k=0}^{t-1}L_T^{\,t-1-k}r_k^2.
    \label{eq:app_timewise_bound}
\end{equation}
Multiply by $w_t\geq0$, sum over $t=1,\dots,T$, and exchange the order of summation:
\[
    \sum_{t=1}^Tw_tS_t\sum_{k=0}^{t-1}L_T^{\,t-1-k}r_k^2
    =\sum_{k=0}^{T-1}\beta_kr_k^2,
\]
with $\beta_k$ as in \eqref{eq:propagation_weight}. This gives \eqref{eq:rollout_training_bound}.
\end{proof}

The inequality holds trajectory-wise, so it also holds in expectation over any distribution of trajectories for which the expectations are finite.

\section{Implement Details}
\label{app:impl-details}

\paragraph{Data.}
We train \method{} on MOVi-Spheres~\citep{HOPNet2025,greff2022kubric} and our synthesis dataset.
MOVi-Spheres contains 1200 scenes with 3 to 10 spheres of varying radius, mass, restitution, and friction, simulated at 1200\,Hz under gravity $10\,\mathrm{m/s^2}$ and stored at 240\,Hz for 2\,s. Scenes are split 80/10/10 into training, validation, and test sets with a fixed seed. We subsample the trajectories to 60\,Hz and cut them into non-overlapping windows of 21 frames which aligns with the video generator~\citep{jiang2025vace}: the operator receives the first frame and predicts the next $K=20$. A pair or floor contact label is positive for a transition when the minimum gap over the 240\,Hz frames between the two sampled frames is at most $0.02$.
Coordinates are mapped from the scene box $[-5,5]^2\times[0,5]$ to $[-1,1]^3$.
Each object state $\mathbf{s}^{i}=(\mathbf{p}^{i},\mathbf{v}^{i},\boldsymbol{\omega}^{i},r^{i})\in\mathbb{R}^{10}$ stores the normalized position, the velocity as a normalized displacement per frame, the angular velocity multiplied by the frame interval, and the radius normalized by the mean box extent. Mass, restitution, and friction are read from the scene metadata and enter only the collision operator.

\begin{figure*}[t]
  \centering
  \includegraphics[width=\linewidth]{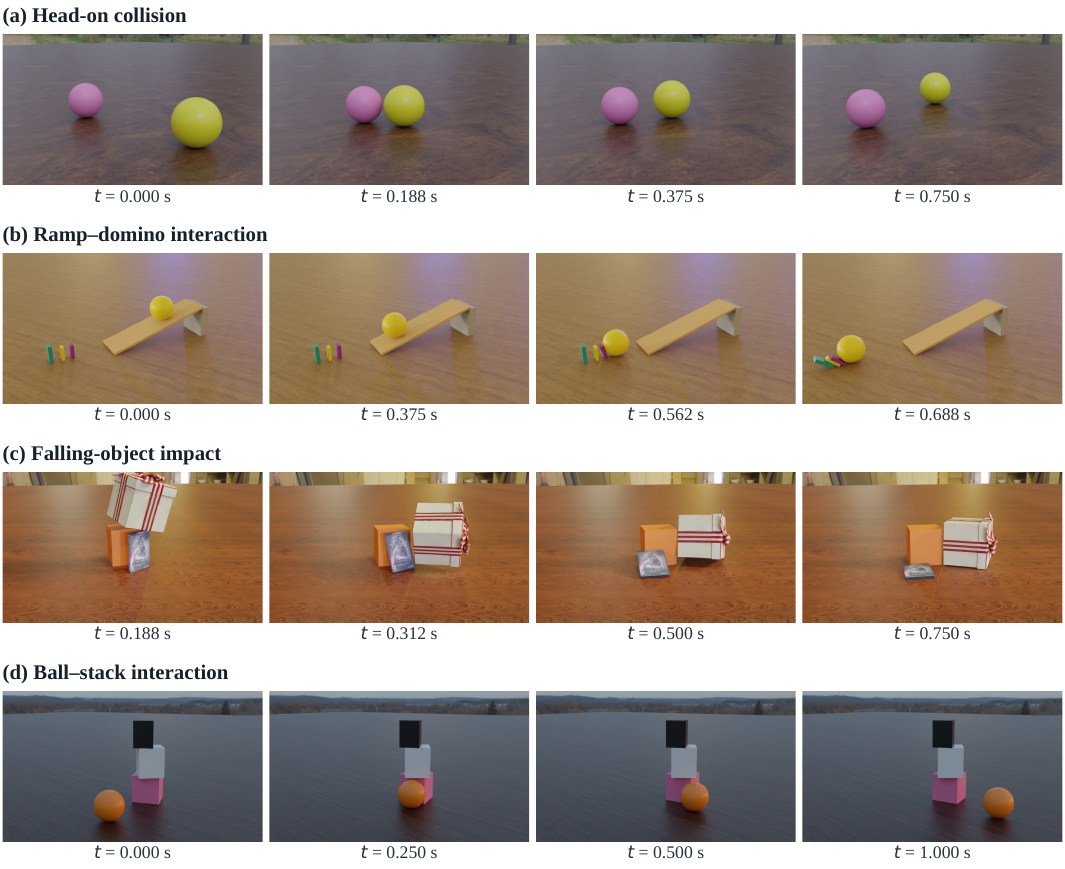}
  \caption{Representative sequences from our synthetic physical interaction dataset. Rows show (a) a head-on collision between two balls, (b) a ball rolling down a ramp toward a row of dominoes, (c) a scanned object falling onto a cluster of objects, and (d) a rolling ball interacting with stacked blocks. Each row follows one reference scene over time; timestamps are relative to the start of its main sequence. Frames are sampled at scene-specific times and retain the full original field of view. The examples illustrate scene diversity and are not a quantitative evaluation.}
  \label{fig:dataset-example}
\end{figure*}

We also construct a synthetic dataset of rigid-body interactions, comprising $6,000$ video sequences organized into $1,500$ scene groups. Some examples are shown in Fig~\ref{fig:dataset-example}. Each scene contains $2–6$ dynamic objects, including balls, blocks, dominoes, and scanned objects drawn from $318$ Google Scanned Objects assets. The dataset covers four interaction families: tabletop rolling ($2,500$ sequences), ramp interactions ($2,164$), falling-object impacts ($668$), and stack interactions ($668$). These families encompass eight scenarios: head-on collisions, passing motions, collisions with object rows, multiple moving objects, single-ramp interactions, opposing-ramp collisions, impacts on object clusters, and impacts on stacked blocks. 

Each scene group contains four variants that share object identities, geometry, materials, lighting, and camera parameters while varying the initial conditions. 
One variant retains the randomly sampled reference conditions; the remaining three use values at the 20\%, 50\%, and 80\% positions within predefined parameter intervals. Depending on the scenario, these variations modify the drop height, rolling speed, or release position along a ramp. Rolling variants also adjust angular velocities and starting positions consistently with the preroll. The camera is determined from the reference setup and held fixed across all four variants. This construction supports controlled comparisons of how changes in initial conditions affect subsequent motion. Sequences are retained without filtering by interaction outcomes.

We generate the sequences using Kubric~\citep{greff2022kubric} with PyBullet simulation and Blender rendering. Each main sequence contains 81 frames at 16 fps and a resolution of $832 \times 480$, spanning five seconds between the first and last frames. Each sample additionally includes a 15-frame preroll at $30$ fps, an initial image for the main sequence, an English scene description, depth maps, and instance segmentation masks. Ground-truth object positions, orientations, linear velocities, and angular velocities are recorded at 240 Hz, together with camera and physical parameters. We partition the dataset by scene group into $1,200$ training, $150$ validation, and $150$ test groups, corresponding to $4,800$, $600$, and $600$ sequences, respectively. All four variants of a group remain in the same partition, preventing variants of the same base scene from appearing across splits.

\paragraph{Data Preprocess}

For input image as prompt, we reconstruct \textbf{executable scenes} with a scene reconstruction agent~\citep{openai2026gpt6astra}. For each case, a case-specific \texttt{gpt-6-astra} session receives the input image and the original text. Then the agent proposes a structured plan and a Blender program. Next we export executable scene information from these programs, including visual meshes, object IDs, initial poses, camera parameters and material property. For each object, we retain the per-frame center estimates together with their actual presentation timestamps, visibility, and ambiguity flags.

\paragraph{Point Sampling.}
The encoder input $\mathbf{P}$ contains $N_p=2048$ points drawn uniformly inside the spheres, with the same number of points for every object regardless of its size. The query set $\mathbf{Q}$ contains $N_q=2048$ points. Half are uniform over the scene box. The other half lie near objects: half of these inside the object volume and half on its surface, perturbed by isotropic Gaussian noise with standard deviation $\sigma_s=0.05$. Each frame is sampled independently. The autoencoder is pretrained on space--time segments of 16 frames with the time coordinate normalized to $[-1,1]$; for a single frame, all points carry $t=0$.

\subsection{Architecture}

\paragraph{Occupancy Autoencoder.}

\begin{figure}[ht]
  \centering
  \includegraphics[width=\linewidth, height=7cm]{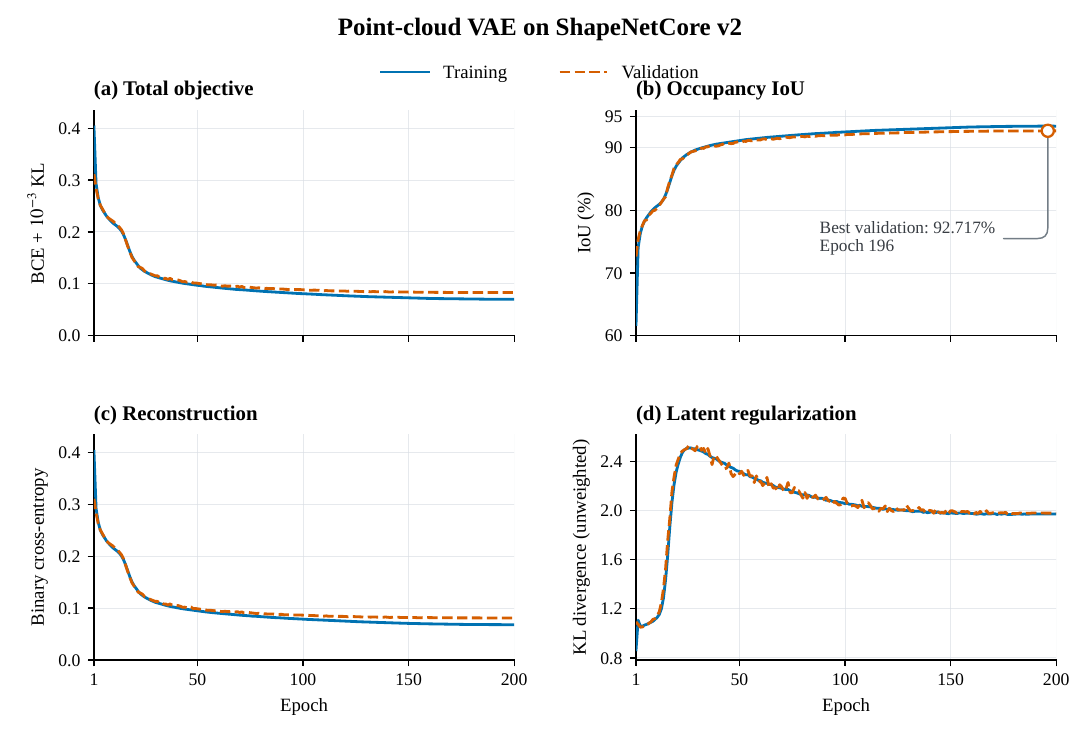}
  \caption{Training curves of the point-cloud-input occupancy VAE (OccVAE3D) on ShapeNetCore v2. (a) Total objective, defined as binary cross-entropy plus $10^{-3}$ times the KL divergence. (b) Occupancy intersection-over-union, pooled over the sampled query points. (c) Reconstruction binary cross-entropy. (d) Unweighted KL divergence, averaged over latent elements. The highest validation IoU is 92.717\% at epoch 196.}
  \label{fig:app_pointcloudvae}
\end{figure}

Fourier features with eight octaves map each four-dimensional point to width 256. The encoder uses one cross-attention block from 256 learned query tokens to the points, followed by eight pre-norm self-attention blocks with eight heads and feed-forward ratio 4. Two linear heads produce $\boldsymbol{\mu}$ and $\log\boldsymbol{\sigma}^2$ of size $256\times32$. The decoder projects the tokens to width 256, applies eight self-attention blocks, and lets each query cross-attend to the result before an MLP predicts its occupancy logit. The reported autoencoder is trained with KL weight $\beta_{\mathrm{KL}}=0.001$. The training curves are shown in Fig.~\ref{fig:app_pointcloudvae}.

\paragraph{Latent neural operator.}
Two-layer GELU MLPs~\citep{hendrycks2023gelu} project the reference coordinates $\boldsymbol{\xi}$, evenly spaced in $[-1,1]$, and the layer-normalized slot features to width 128. The physics cross-attention computes $\mathbf{W}\in\mathbb{R}^{256\times128}$ from the projected coordinates. A softmax over slots aggregates the 256 slots into 128 modes, eight attention blocks with four heads act on the modes, and a softmax over modes distributes the result back to the slots, where a two-layer MLP with a zero-initialized output layer returns the latent update. Since $\boldsymbol{\xi}$ is shared by all samples, $\mathbf{W}$ does not depend on the input.

\paragraph{Free-motion Proposal.}
State adapter $\mathcal{A}_{\mathrm{c}}$ projects each object state with a two-layer MLP and the latent slots with a linear layer to width 128, applies one four-head cross-attention from slots to objects with absent objects masked, and maps the normalized sum back to the latent width as a residual. The proposal head $\mathcal{P}_{\mathrm{free}}$ adds an eight-band Fourier embedding of the target time to each object token and applies cross-attention to $\widehat{\mathbf{z}}_{k+1}$, masked self-attention among objects, and a feed-forward block, each followed by a residual connection and layer normalization, all at width 128 with four heads. Its zero-initialized linear output gives the corrections of free-motion proposal, bounded by $0.05$ for position and $0.25$ for linear and angular velocity in normalized units. The radius is copied from the initial state. The collision operator does not change $\boldsymbol{\omega}$, so spin changes from friction are learned only by this head.

\paragraph{Surface operator.}
Each object is represented by 128 oriented surface samples placed on a Fibonacci lattice. For every object, the eight nearest samples of the other objects give a geometric gap and normal. Their relative positions and normals, scaled by the object radius, are encoded by a shared MLP of width 128, pooled by learned attention weights, and mapped to a gap correction bounded by $0.25$ and a normal correction bounded by $0.5$; both output layers are initialized to zero. The operator is pretrained on exact gap and normal targets with a smooth-$L_1$ gap loss weighted towards near and contacting queries and a cosine normal loss, reaching contact F1 $0.934$ against $0.840$ for the uncorrected estimate on validation windows. It is frozen during Stage~2 training.

\paragraph{Collision operator.}
Contact gaps and normals for the proposed positions are computed in closed form for sphere pairs and against the floor plane; the surface-operator estimates enter as features. Each contact edge is described by 21 features: the gap, the normal, the relative velocity and its normal and tangential speeds, both radii, both inverse masses and the reduced mass, the restitution and friction priors, the surface-operator gaps and their alignment with the contact normal, and an indicator for the static floor. A two-layer GELU edge network of width 128, shared by all edges, encodes them, and zero-initialized linear heads give $\rho_{ij}$ and the coefficient corrections, so the operator starts as exact geometry with the material priors of the data. The contact probability uses tolerance $\epsilon=0.02$ and temperature $\tau_{\mathrm{c}}=0.02$. The restitution and friction priors are the geometric means of the per-object values for body pairs and the object's own value against the floor; the learned corrections enter through $0.5\tanh(\cdot)$, restitution is clipped to $[0,1]$, and friction to $[0,2]$. The velocity correction of each object is clipped to norm $50$, which leaves momentum conservation exact whenever the clip is inactive. Overlaps are removed by moving each body along the normal by half of the penetration depth, weighted by its share of the inverse mass, and positions additionally follow half of the velocity correction over one frame.

\subsection{Training}
\label{app:impl-training}

\paragraph{Loss weights.}
The occupancy temperature is $\tau_{\mathrm{o}}=0.02$, and the state losses use smooth-$L_1$ with threshold $\beta_{\mathrm{SL1}}=0.02$. The weights of Eq.~\ref{eq:latent-dynamics-loss} are $\lambda_z=\lambda_o=1$, $\lambda_s=1$ with equal weights on position, velocity, and angular velocity, and $\lambda_c=0.2$ for the averaged object and pair contact terms and, separately, for the floor contact term. All contact terms use focal parameters $\alpha=0.75$ and $\gamma=2$.

\paragraph{Optimization.}
All stages use AdamW with weight decay $10^{-4}$ and gradient clipping at
norm $1$. The autoencoder is trained for 200 epochs with learning rate
$4\times10^{-5}$ and batch size 64. The surface operator is trained with
learning rate $3\times10^{-4}$, batch size 64, and a cosine schedule on 6000
training windows, with early stopping after at least 8 epochs. The dynamics
stage uses learning rate $10^{-4}$ with a cosine schedule and a total batch
size of 128 for 100 epochs, and the checkpoint with the best validation
occupancy IoU is kept. Training runs with data parallelism on eight GPUs.

\section{Additional Experiments}
\label{app:additional-experiments}

\paragraph{Metrics and time axis.}
The test set contains 1,060 valid object-times and 53 valid final objects. ADE averages position error over valid object-times; FDE averages over valid objects at the final prediction. Each error is normalized by the object's equivalent radius $R_{\mathrm{eq}}$, clamped below at 1\,mm.
The saved 20 prediction errors correspond to \texttt{source\_frames[1:]}: subtracting the initial source frame and dividing by the 240\,Hz source rate gives times $1/60,\ldots,20/60$\,s. No zero-error point is inserted at $t=0$.

\begin{figure}[t]
    \centering
    \includegraphics[width=\linewidth]{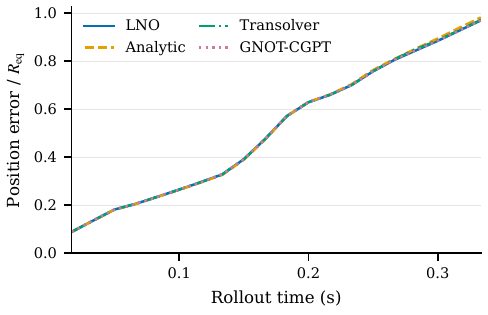}
    \caption{\textbf{Early state-error accumulation.}
    Saved radius-normalized position errors, averaged over valid objects at
    each prediction time on the same 16 test scenes. The vertical axis starts
    at zero. This is one training seed and a 20-step ($\simeq0.333$\,s)
    rollout under the shared latent-slot interface, not a convergence or
    long-horizon video comparison.}
    \label{fig:processor-pilot}
\end{figure}

\begin{figure}[t]
  \centering
  \includegraphics[width=\linewidth]{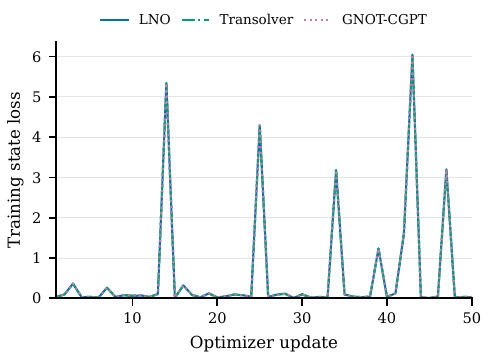}
  \caption{\textbf{Actual loss over the 50 training updates.}
  Each update uses its prescribed training window; all three learned arms
  see the same sequence. These raw state losses mix changing window
  difficulty with optimization, and are not fixed-validation convergence
  curves. The analytic arm is not trained.}
  \label{fig:processor-pilot-loss}
\end{figure}

\paragraph{Scope.}
The study probes early optimization under a matched update budget and similar active processor capacity. A single seed, 50 updates, and this short horizon do not establish a production-model ranking, statistical significance, video-control gains, or long-time generalization.

\end{document}